\documentclass[10pt,twocolumn,letterpaper]{article}

\usepackage[top=.75in,bottom=.75in,left=.75in,right=.75in]{geometry}
\usepackage{sectsty}
\sectionfont{\large}
\subsectionfont{\large}
\subsubsectionfont{\large}
\paragraphfont{\large}

\usepackage{times}
\usepackage{graphicx}
\usepackage{amsmath,amsthm}
\usepackage{amssymb,empheq}
\usepackage{latexsym}
\usepackage{epstopdf}
\usepackage{url}
\usepackage{multirow}

\usepackage[bf,font=footnotesize]{caption}
\usepackage[labelfont=bf,font=scriptsize]{subcaption}

\usepackage{algorithm}
\usepackage{algorithmic}

\usepackage[usenames]{color}
\definecolor{shadecolour}{gray}{0.4}

\graphicspath{{./}{./Fig/}{./fig/}{./Figs/}{./figs/}}

\newcommand{\BY}{{\mathbf{Y}}}
\newcommand{\BX}{{\mathbf{X}}}
\newcommand{\BD}{{\mathbf{D}}}

\newcommand{\BW}{{\mathbf{W}}}

\newcommand{\Bf}{{\mathbf{f}}}

\newcommand{\Bz}{{\mathbf{z}}}

\newcommand{\BI}{{\mathbf{I}}}
\newcommand{\T}{{\!\top}}
\newcommand{\Bx}{{\mathbf{x}}}

\renewcommand{\Lambda}{\varLambda}

\newcommand{\st}{{\,\,\mathrm{s.t.\,\,}}}

\ifx\theorem\undefined
\newtheorem{theorem}{Theorem}
\newenvironment{theorem*}{\par\noindent{\bf Theorem\ }}{\hfill\\[2mm]}

\newenvironment{corollary*}{\par\noindent{\bf Corollary\ }}{\hfill\\[2mm]}

\fi

\newcommand{\Bc}{{\mathbf{c}}}

\usepackage[pagebackref=true,breaklinks=true,letterpaper=true,colorlinks,bookmarks=false]{hyperref}

\usepackage{cite}
\usepackage{eucal,bibspacing}

\date{}
\author{Fumin Shen\thanks{
   Part of this work was done
   when the first author was visiting The University of Adelaide.
   }, ~
   Chunhua Shen\thanks{Correspondence should be addressed
   to C. Shen.}
   \\
$^*$ Nanjing University of Science and Technology, China
~ ~ ~ ~
$^\dag$ The University of Adelaide, Australia
}

\begin{document}

\title{Generic Image Classification Approaches Excel on Face Recognition}

\maketitle
\thispagestyle{empty}

\begin{abstract}

The main finding of this work is that the standard image classification pipeline,
    which consists of dictionary learning, feature encoding,
spatial pyramid pooling and linear
classification, {\em outperforms  all state-of-the-art face recognition methods} on the tested
benchmark datasets (we have tested on AR, Extended Yale B,  the challenging  FERET, and LFW-a datasets).
This surprising
and prominent result suggests that those advances in generic image classification can be
directly applied to improve face recognition systems. In other words, face recognition may not need to be
viewed as a separate object classification problem.

While recently a large body of residual based face recognition methods focus on developing
complex dictionary learning algorithms, in this work we show that a dictionary of
randomly extracted patches (even from non-face images) can achieve very promising results using the
image classification pipeline. That means, the choice of dictionary learning methods may not be
important. Instead, we find that learning multiple dictionaries using different low-level
image features often improve the final classification accuracy.
Our proposed face recognition approach offers the best reported results on the widely-used
face recognition benchmark datasets. In particular, on the challenging FERET and LFW-a datasets,
     we improve the best reported accuracies in the literature by about 20\% and 30\%
respectively.

\end{abstract}

\section{Introduction}
\label{SEC:Intro}

In recent years, researchers have spent significant effort on appearance based face recognition.
In particular, (sparse) representation based face classification has shown success in the
literature \cite{Wright09,CRCzhanglei2011}.
Given a test face image, the classification rule is based on the minimum representation error
over a set of training facial images. In this category, one of the well-known methods might be the sparse
representation based classifier (SRC) \cite{Wright09}. SRC  linearly represents a
probe image by all the training images under the sparsity constraint/regularisation using the $L_1$ norm. The
success of SRC has induced a few sparse representation based algorithms
\cite{RSC11,ESRC2012,yang2010gabor,dengdefense2013}.
Competitive results have been observed using non-sparse $ L_2 $ norm regularised representations
\cite{CRCzhanglei2011,LRC10,Javen2011}. The main advantage of these approaches is its computational
efficiency due to the closed-form solution.
Zhang et al.\ argued that it is  the collaborative representation but not the $L_1$  sparsity  that
boosts the face recognition performance \cite{CRCzhanglei2011}.
Their method is dubbed as Collaborative Representation Classification (CRC).

   Instead of focusing on the encoding strategy and directly using the training samples
   as the dictionary,  another research topic is  to learn a dictionary using improved variants of sparse coding.
   Yang et al.\ \cite{yang2011fisher} proposed a dictionary learning method using the Fisher discrimination criterion.
   Dictionary learning was formulated as an extended K-SVD problem in
   \cite{zhang2010discriminative} and \cite{jiang2011learning}, where the classification
   error was taken into the objective function.
   To alleviate the effect of image noise contamination, low-rank minimisation of dictionary
   was utilised with the help of class discrimination \cite{ma2012sparse} or structural
   incoherence of basis \cite{chen2012low}.

   Almost all of these methods use holistic images. Local feature descriptors such as
   histograms of Local Binary Patterns (LBP) \cite{FR_LBP06}, Gabor wavelets \cite{GaborWavelet93}
   have been proven to improve the robustness of face recognition systems.
   A {\em heuristic} method is the modular approach, which first partitions the entire facial
   image into several blocks and then classification is made independently
   on each of these local blocks. The intermediate results on local patches are finally
   aggregated, e.g., by distance-based evidence fusion (DEF) \cite{LRC10} or majority voting
   \cite{Wright09}. Zhu et al.~\cite{zhu2012multi} proposed a multi-scale patch
   based method, which fused the results of patches of all scales with the scale weights
   learned by regularized boosting.

   {\em All the aforementioned representation based methods make classification decisions based on the
   representation errors/residuals}. In contrast, the encoded coefficients
   themselves have not been fully exploited for face recognition.
   This is very different from the simplest, standard pipeline of generic image classification,
   such as the Bag-of-Features (BoF) model.
   \cite{coates2011importance}.
   The BoF model
   \cite{LLC2010,coates2011importance} is generally composed of dictionary learning
   on local features (either raw pixels or dense SIFT features),
   feature encoding and spatial pyramid pooling. The locally encoded and
   spatially aggregated feature representation, coupled with an efficient linear classifier (e.g.,
   linear support vector machines (SVM)), has been the standard approach to generic image classification,
   including fine-grained image classification, and object detection.

   A striking and surprising result of our work here is that
   {\em the simple,  standard generic image classification approach
   using unsupervised feature learning outperforms the best face recognition methods.}
   This finding suggests that face recognition may not need to be tackled as a separate problem---generic object
   recognition approaches work extremely well on face recognition.
   A bold suggestion is that for the problem of face recognition, it may be better to
   shift the research focus from the recognition component to the pre-processing component, e.g.,
   facial feature detection and alignment, which often benefits the subsequent recogniser.

\begin{figure*}[t!]
\centering
\includegraphics[width = 0.65\textwidth]{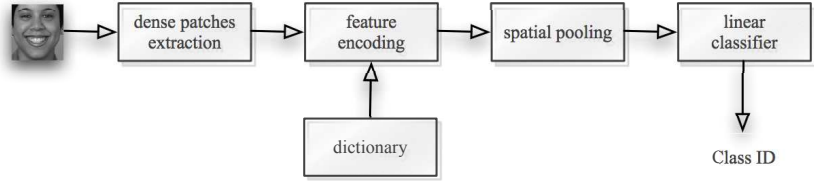}
\caption{An illustration of the standard image classification pipeline.}
\label{Fig:chart}
\end{figure*}

   Recently, Coates et al.\ showed that for  image classification, a very simple module combination
   (dictionary of patches learned using K-means or even random selection with a simplified non-linear
   encoder) can achieve state-of-the-art image classification performances on benchmark datasets
   such as CIFAR and CalTech101
   \cite{coates2011importance,coates2011analysis}.
   Our baseline face recognition system follows this simple pipeline of
   \cite{yang2009linear,coates2011importance},
   which is illustrated in Fig.\ \ref{Fig:chart}.
  Namely, we solve the face recognition problem using the pipeline of
  dictionary learning, feature encoding and spatial pooling followed by linear classification.
  We also thoroughly evaluate the impact of each stage in the pipeline on the final face
  recognition performance. Some interesting results are found, which provide insightful
  suggestions to the face recognition community.

  In summary, our main contributions are as follows.
\begin{itemize}

\setlength{\itemsep}{-1.2mm}

\item[1.]

   We show that, given the raw-pixel face images, simple feature learning coupled with a highly
efficient linear classifier can achieve state-of-the-art face recognition performances on
widely-used face benchmark datasets. We thus suggest that face recognition can be tackled
as an instance of generic image classification, advances of which can help boost
the face recognition research.

\item[2.]
The generic image classification pipeline (including dictionary learning, encoding and
pooling) is thoroughly evaluated in the scenario of face recognition.
We show that dictionary learning is less important than that of
feature encoding. This is consistent with the observation in \cite{coates2011importance}.
   In particular, we show that a dictionary using random patches, without learning involved,
   can work very well.

   Another interesting finding is that, using this recognition pipeline, a dictionary learned
   using {\em non-face patches} can also achieve excellent results.
   That is, a face image patch can be represented by using non-face image patches.
   In contrast, all the previous representation based face recognition methods (SRC \cite{Wright09},
   CRC \cite{CRCzhanglei2011}, LRC \cite{LRC10} etc.) are restricted to use facial
   training images to form the dictionary.

\item[3.]
Composed of inner product and simple non-linear activation function, the soft
threshold encoding performs close to (however much faster than) sparse coding. Actually in
some situations (e.g., with large class number and less training data) soft threshold
achieves even better results than sparse coding.
This deviates the results found on the CIFAR and CalTech101
image classification \cite{coates2011importance}.

\item[4.]
Last, we show that a simple fusion of the learned features using raw-pixel intensity and
LBP  further improves the classification performance over features learned using raw-pixel or LBP alone.

\end{itemize}

   Next, before we present our main results, we briefly introduce the image classification pipeline using
   unsupervised feature learning.

\section{Image classification and feature learning}

Image classification, which classifies an image into one or a few categories, has been one
of the most fundamental problems in computer vision.
Possibly due to the special characteristics of face images and historical reasons,
face recognition has so far been considered as a different task from generic image classification.
However, in essence, face recognition is a sub-category or fine-grained object classification problem.
Indeed, for the first time, we show that generic object recognition methods work extremely well on face
recognition.

The generic image classification pipeline (shown in Fig.~\ref{Fig:chart}) has achieved
state-of-the-art performances \cite{LLC2010,coates2011importance}. Low-level features on
local patches are usually densely extracted and pre-processed (normalisation and whitening)
in the first step. It has
been shown that the pre-processing steps can have considerable influence on the final classification performance
 \cite{coates2011analysis}.
   With the extracted local patches, an over-complete dictionary is formed by using unsupervised
learning, e.g, K-means, K-SVD, or sparse coding \cite{yang2009linear}.
The image patches are then encoded with the learned dictionary, e.g., using the hard vector
quantisation, variants of sparse encoding \cite{yang2009linear,LLC2010} or soft threshold
\cite{coates2011importance}. To encode the spatial information as well as reduce the dimension of the generated features,
   the encoded features are pooled over the
pre-defined spatial cells either using average pooling or max-pooling, results of which are
concatenated to form the final representation of an image. Last, a
linear classifier  is trained with the learned image features.

\subsection{Unsupervised dictionary learning}

In this work, we use the following methods to construct the dictionary $\BD \in
\mathbf{R}^{d \times m}$. Here  $m$ is the number of atoms, and $ d $ is the input dimension.
   We have chosen these methods  mainly due to their computational scalability and efficiency.
Suppose that we have extracted $N $ small local patches
$\{\Bx_i\}$ from the training data.
\begin{itemize}
\item[1.] \textbf{Random selection:}
The dictionary $\BD$ is directly formed by the $m$ patches randomly selected from $\{\Bx_i\}$.
No learning is involved.
\item[2.] \textbf{K-means clustering:}
   We apply K-means clustering on  local patches $\{\Bx_i\}$ and
   fill the columns of $\BD$ with the $m$ cluster centres.
\item[3.]  \textbf{Sparse coding:}
Sparse coding forms  an over-complete dictionary by minimising the reconstruction error
with the $L_1$ sparsity constraint. The objective function can be written as
\cite{coates2011importance}:
\begin{align}
\min_{D, \Bf_i} & \sum_i \|\BD\Bf_i - \Bx_i\|^2_2 + \lambda \|\Bf_i\|_1 \notag\\
\st & \|\BD_j\|_2 = 1, \quad \forall j.
\label{EQ:SC}
\end{align}
Here $\BD_j$ is the $j$th dictionary atom ($j$th column of the matrix $ \BD $).

\item[4.] \textbf{K-SVD:}
   As a generalization of K-means,
   K-SVD  shares a similar objective function as sparse coding \cite{KSVD2006}.
   It has been  used for generating dictionary for residual based face recognition
   \cite{zhang2010discriminative,jiang2011learning}.
\end{itemize}

\subsection{Feature encoding}

     Once the dictionary is obtained,
     local patches are then fed into the feature encoder to generate
     a set of codes $\{\Bf_i\}$.
     We use the following encoding methods.
\begin{itemize}
\item[1.] \textbf{Sparse coding: } With $\BD$ fixed in \eqref{EQ:SC},
   $\Bf_i$ can be obtained by solving the LASSO problem.
\item[2.] \textbf{Locality-constrained linear coding:}
Compared with sparse encoding, LLC \cite{LLC2010} is more efficient,
         which solves for the code of a sample by a constrained least square problem with
         its $K$ nearest neighbours.
\item[3.] \textbf{Ridge regression:}
If one replaces the $ L_1 $-norm regularisation with the $ L_2 $-norm regularisation,
   one has the ridge regression problem, which admits a closed-form solution.
   We can generate the code directly by the ridge regression problem.

   In \cite{CRCzhanglei2011}, the authors reformulate the $ L_1 $
   optimisation of SRC into a $ L_2 $ ridge regression problem for face recognition.

\item[4.] \textbf{Soft threshold:} With a fixed threshold $\alpha$,
   this simple feed-forward non-linear encoder writes \cite{coates2011importance}:
\begin{align}
\Bf_j(\Bx) =& \max\big\{0, \BD_j^\T\Bx - \alpha \big\},\\
\Bf_{j+m}(\Bx) = &\max\big\{0, -\BD_j^\T\Bx - \alpha \big\}.
\label{EQ:ST}
\end{align}
Here $\Bf_j$ is the $j$th entry of the encoded feature vector $\Bf$.
\item[5.] \textbf{K-means triangle:} As a `softer' extension of the hard-assignment coding, with the $m$ centroids $\{\Bc_j\}$ learned by K-means, K-means triangle encode $\Bx$ as:
\begin{equation}
\Bf_j(\Bx) = \max\big\{0, \mu(\Bz) - \Bz_j\big\},
\label{EQ:KT}
\end{equation}
where $\Bz_j = \|\Bx - \Bc_j\|$ and $\mu(\Bz)$ is the mean of $\Bz$.

We can also use this encoding method by replacing $\{\Bc_j\}$ with the
bases generated by other dictionary learning methods other than  K-means.
See \cite{coates2011analysis}
   for details.
\end{itemize}

\subsection{Linear classifiers}
\label{SEC:classifier}

A simple ridge regression based classifier is used in  \cite{gong2011comparing}.
We use the same linear classifier for its computational efficiency.
One can of course use linear SVM such as LIBLINEAR  \cite{liblinear08}.
However, for large-scale high-dimensional data, LIBLINEAR is still slow.
The ridge regression classifier has a closed-form solution, which is very fast.
Despite its simplicity, the classification performance of this ridge regression approach
is on par with linear SVM \cite{gong2011comparing}.

Let $\BX \in \mathbf{R}^{n \times d}$ be the training data of $n$ samples of dimension
$d$ and $\BY \in \{0, 1\}^{n \times c}$ the label matrix: $\BY_{ij}=1$ if data point $i$
belongs to class $j$ of total $c$ classes and 0 otherwise.
Unlike SVM, this classifier generates the classifier matrix with a closed-form solution
$\BW = (\BX^{\T}\BX + \delta \BI)^{-1}\BX^{\T}\BY$, where $\BI$ is the identity matrix.
Then an image $\Bz$ is classified by the maximum of $\BW^{\T}\Bz$.
Another benefit of this classifier is that, compared to the one-versus-all SVM, it only needs
to compute $\BW$ once. When the number of classes is large, training of multiple one-versus-all classifiers also
creates the problem of imbalanced data (the number of positive data is much smaller than that of negative data).
This ridge regression classifier does not have this problem.
In practice the dimension $d$ could be over hundreds of thousands when with several
pooling pyramid levels.

We thus compute the matrix $\BW$ by its equivalent $\BW = \BX^{\T}(\BX\BX^{\T} +
\delta \BI)^{-1}\BY$, where it only needs to invert much smaller $n \times n$ matrix\footnote{The
   parameter $\delta$ is empirically set to $0.005$ in all of our experiments. Careful tune of this
parameter might slightly improve the performance.}.

\section{Evaluation of the pipeline for face recognition}

In this section, we  test the feasibility and performance of classifying face images exactly same as
 in classifying natural images, and try to discover what the most important component is.
Thorough cross evaluations of the components in the image classification pipeline are
conducted on face recognition.

Following  \cite{coates2011importance},
 local patches are extracted on data with patch size of $6 \times 6$ pixels and a stride of $1$
pixel, followed by contrast normalization and ZCA whitening \cite{ICA2000}. It has been
proven that dense feature extraction and the pre-processing step are critical for achieving
better performance.

Unless otherwise specified, we generate the dictionary from 50,000 randomly extracted patches on the training data.
In all the experiments except those in section \ref{SEC:dict}, the threshold $\alpha$ in soft threshold and the regularization
parameter $\lambda$ in sparse coding are set to 0.25 and 1, respectively.
As a default setup in the evaluation, the final features are formed by max pooling over 3 pyramid levels.

We mainly conduct evaluations on the datasets of AR and Extended Yale B, and LFW. The datasets
description can be found in section \ref{SEC:Exp}. In this section we use 13 and 32
training samples each class for AR and Extended Yale B, respectively. The rest are used
for testing. For LFW-a, 5 samples are randomly
selected for training and another 2 samples for testing in each class.

\subsection{Dictionary learning and feature encoding}
\label{SEC:dict}

\begin{figure}[]
\centering
\includegraphics[width=0.45\textwidth]{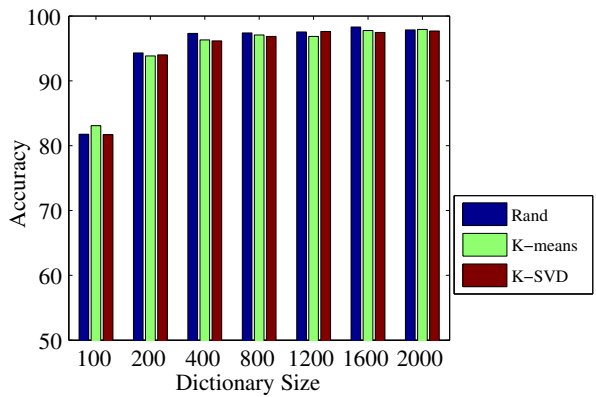}
\caption{Impact of the size of the dictionary (obtained by random selection, K-means
   and K-SVD) on the recognition accuracy (\%). The AR dataset is used.
   Here the encoding strategy is soft
   thresholding for efficiency.}
\label{Fig:DictionarySize}
\end{figure}

First, we evaluate the impact of the dictionary size on face recognition performance.
Fig.~\ref{Fig:DictionarySize} shows the results of our method with the dictionary size
varying from 100 to 2,000. It is not surprising to see that {\em the accuracy is consistently
improved as the dictionary size increases.} Coates et al.\ \cite{coates2011importance} observed
that a large dictionary often leads to improved classification accuracy, especially
when one has a limited number of labelled training data for training the final classifier.
     This is the case for face recognition.
In the following experiments we set the dictionary size to 1,600.

Next, we then examine the importance of dictionary learning and feature encoding methods for
face recognition. Four methods of dictionary construction: randomly selected patches (uniform sampling),
K-means, K-SVD \cite{KSVD2006}, sparse coding \cite{yang2009linear}; and five encoding
algorithms: sparse coding, LLC \cite{LLC2010}, $ L_2 $ ridge regression, soft threshold, K-means
triangle \cite{coates2011analysis} are  evaluated.
We set the maximum number of iterations for K-SVD to 30,
and  the number of nearest neighbours $K = 5$ for LLC.
We conduct K-SVD with  sparsity level $T = \{3, 5, 10, 15, 30\}$, sparse coding with $\lambda = \{0.25, 0.5, 1.0, 1.5\}$, LLC with the regularization parameter $\delta = \{0.001, 0.01, 0.1, 1\}$, soft threshold with $\alpha = \{0.1, 0.25, 0.5, 0.75, 0.1\}$.
By maximizing over these parameters, best cross validation results with each combination of the dictionary learning and encoding methods are reported.
Table~\ref{Tab:Dict_Encoder_AR} and Table~\ref{Tab:Dict_Encoder_LFW} list the recognition
results on the AR and LFW datasets, respectively.

\begin{table}[]
\centering
\begin{tabular}{r cccccc}
\hline
Encoder & ST & SC & LLC & RR &  KT & VQ\\
 \hline
Random & 97.8   & 98.8  &  98.2  &  95.7 &   97.5 & 94.5\\
 K-means&    97.8 &  98.5 &  97.2  &  95.6   & 97.1 &92.9\\
K-SVD&     98.2 &   98.6  &  98.4  &  96.1   & 97.8 &95.1\\
 SC &    98.1 &  98.4   & 98.9  &  97.7 &   98.5 &94.3\\
 \hline
\end{tabular}
\caption{ Accuracies (\%) with various dictionary learning methods (shown in the first
column) and encoding methods (shown in the first row) on AR. `SC', `LLC', `RR', `ST',
`KT' and `VQ' represent `Sparse Coding',
   `Locality-constrained Linear Coding',
   `Ridge Regression', `Soft Threshold', `K-means Triangle' and hard `Vector Quantization', respectively.
      {\em The results at each column are similar,
which shows that dictionary learning methods
   do not  have a significant impact on the final performance.
}
}
\label{Tab:Dict_Encoder_AR}
\end{table}

\begin{table*}[]
\centering
\begin{tabular}{r cccccc}
\hline
Encoder & ST & SC & LLC & RR &  KT & VQ\\
 \hline
Random & $75.4 \pm 1.6$   & $76.4 \pm 1.8$  &  $76.5 \pm 0.5$  &  $68.2 \pm 3.6$ &   $78.0 \pm 1.2$ & $62.5 \pm 2.1$\\
 K-means&    $75.4 \pm 1.6$ &  $77.2 \pm 2.3$  & $72.9 \pm 2.2$  &  $64.6 \pm 2.1$   & $77.9 \pm 0.9$ &$60.6 \pm 2.1$\\
K-SVD&     $75.4 \pm 1.6$ &   $77.6 \pm 2.8$  &  $75.1 \pm 0.8$  &  $69.1 \pm 1.4$   & $79.0 \pm 1.2$ &$61.6 \pm 1.1$\\
 SC &    $74.7 \pm 1.9$  &  $74.3 \pm 3.7$   & $75.7 \pm 3.1$  &  $72.6 \pm 1.1$ &   $80.0 \pm 2.3$ &$60.3 \pm 0.4$\\
 \hline
\end{tabular}
\caption{ Accuracies (\%) with various dictionary learning methods (shown in the first
column) and encoding methods (shown in the first row) on LFW. Results are based on 5 independent runs.      
}
\label{Tab:Dict_Encoder_LFW}
\end{table*}

   As we can see, with an encoder fixed (such as soft threshold or sparse coding),
   different dictionary learning methods lead to similar classification results.
   In other words, dictionary learning is less critical in terms of the final performance.
   As observed in \cite{coates2011importance}, for face recognition we see similar observations:
   simple algorithms like K-means can work as well as sparse coding and K-SVD in most
   cases.
   In general, Vector Quantization (VQ) as in the traditional bag-of-visual-words model,
   performs  worse than other encoding methods because  VQ
   may have lost much useful information due to the aggressive hard-assignment
   coding.

   In \cite{coates2011importance}, the authors show that
   a dictionary of randomly selected patches usually performs slightly
   worse on CIFAR and CalTech101.
   While for face recognition, we see that even with a dictionary of
   randomly selected patches, the performance is as good as other dictionary learning methods such as
   sparse coding when the size of the dictionary is large (1,600 is sufficient in our experiments).

   This is in contrast to the representation error based methods, where it is important
   to design an effective dictionary (in order to reconstruct the test image well), e.g., by sample
   variations \cite{ESRC2012,dengdefense2013}, discriminative  learning \cite{yang2011fisher}
   or low-rank minimization \cite{ma2012sparse,chen2012low}.

   Among the encoding approaches, sparse coding  achieves the best results on
   the AR dataset, which marginally outperforms other encoding methods with most dictionary training methods.
   The simple and computationally efficient encoder `soft threshold' and `K-means Triangle' also stably offer
   promising results, which is much better than ridge regression and VQ on this dataset. On the
   relatively challenging dataset LFW, the simple algorithm K-means triangle works even better than all other methods with any dictionary learning methods. Note that K-means triangle does not have a parameter to tune.

\begin{figure*}[]
\centering
\begin{subfigure}{0.23\textwidth}
\centering
\includegraphics[width=0.95\textwidth]{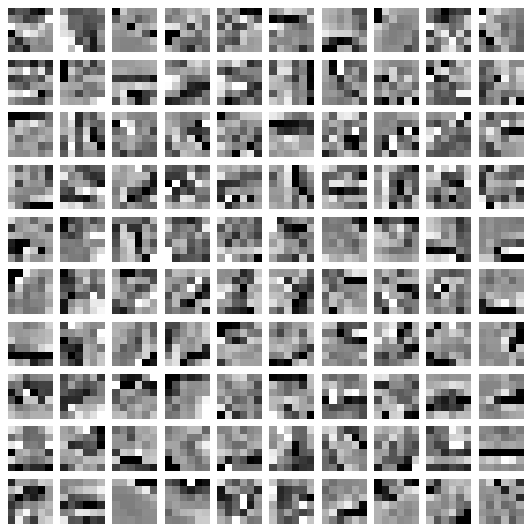}
\caption{Randomly selected patches}
\end{subfigure}
\begin{subfigure}{0.23\textwidth}
\centering
\includegraphics[width=0.95\textwidth]{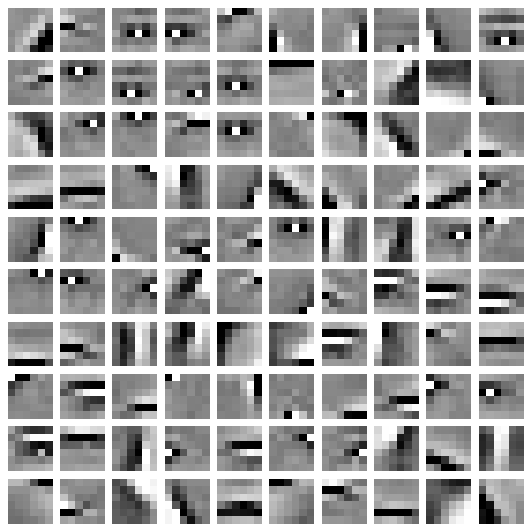}
\caption{K-means}
\end{subfigure}
\begin{subfigure}{0.23\textwidth}
\centering
\includegraphics[width=0.95\textwidth]{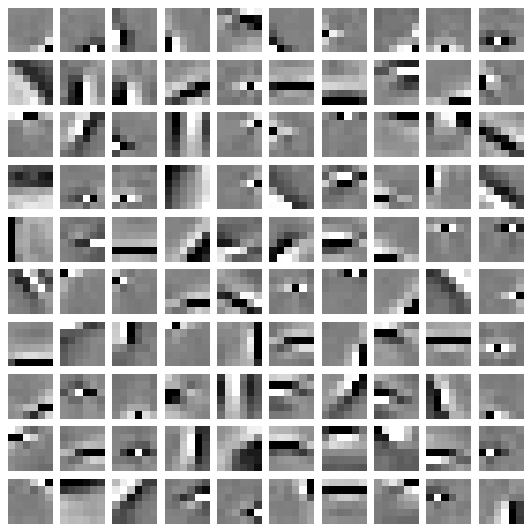}
\caption{K-SVD}
\end{subfigure}
\begin{subfigure}{0.23\textwidth}
\centering
\includegraphics[width=0.95\textwidth]{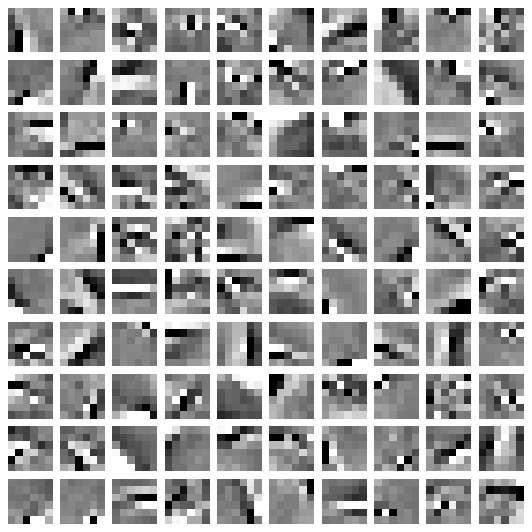}
\caption{Sparse coding}
\end{subfigure}
\caption{Visualisation of some generated dictionary bases by a few different dictionary construction
   methods. The bases generated by different methods appear similarly.}
\label{Fig:dictionary}
\end{figure*}

   Consistent with the finding for generic image classification \cite{coates2011importance}, {\em
   we can see from this experiment that the choice of dictionary learning methods  is less
   important than that of feature encoder in the face image classification pipeline.}
   Random patches combined with a simple encoder, which introduces non-linearity, can work very
   well. In general, the conclusion is that
   most non-linear encoding methods yield promising results for face recognition.
   Here we show some learned bases generated by different dictionary learning methods
   in Fig.\ \ref{Fig:dictionary}.

\begin{figure}[t]
\centering
\includegraphics[width=0.45\textwidth]{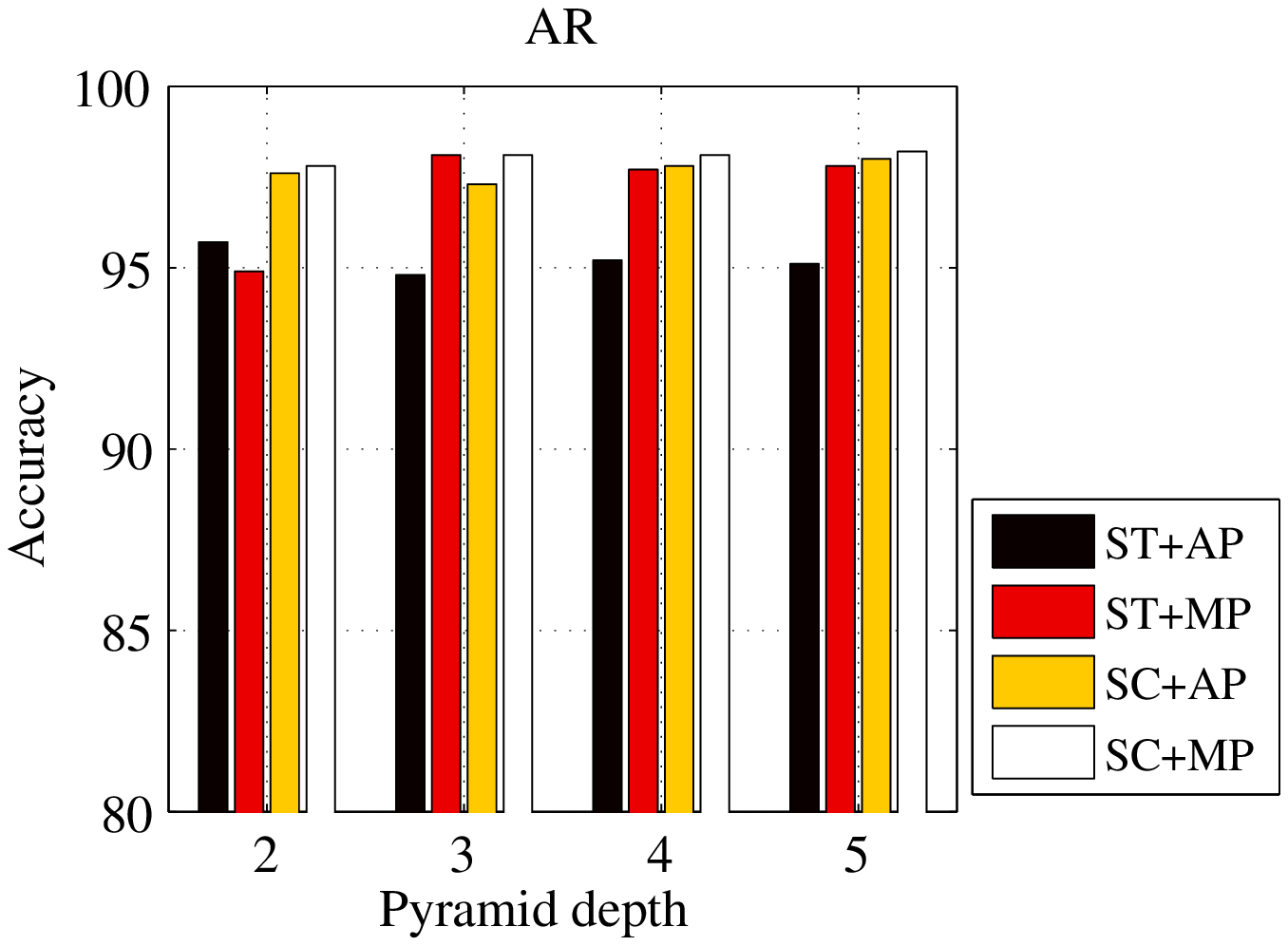}
\includegraphics[width=0.45\textwidth]{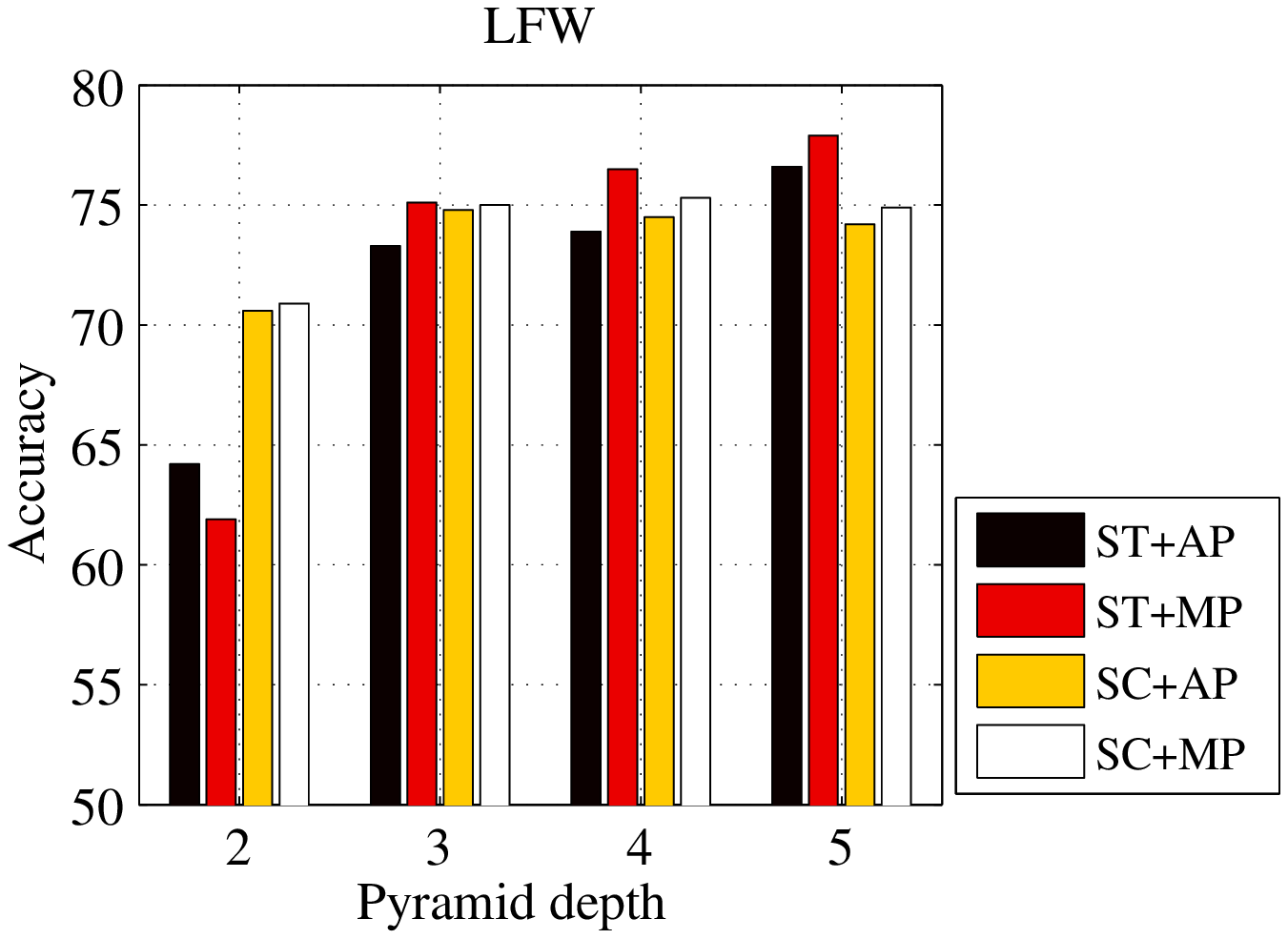}
\caption{Accuracies (\%) with different number of spatial pyramid levels on the
   AR and LFW-a datasets. `ST', `SC', `AP' and `MP' represent `Soft Threshold', `Sparse Coding', `Average Pooling' and `Max Pooling', respectively.}
\label{Fig:Pooling}
\end{figure}

\subsection{Spatial pyramid pooling}

In this subsection, we evaluate the two popular pooling method: sum/average pooling and
max pooling with varying pyramid levels.
With pyramid pooling, the encoded features for all local patches are pooled over each grid of all
spatial pyramid levels \cite{SPM06}. We set the first a few levels to have \{$1 \times 1$,
$2 \times 2$, $4 \times 4$, $6 \times 6$, $8 \times 8$\} equal-sized rectangular pooling
grids.
The pooled features at each level are concatenated to form the final feature vectors.

Fig.~\ref{Fig:Pooling} shows the results on AR and LFW-a \cite{LFW_a}. 
{\em The first observation is that spatial pyramid pooling is important for the final classification
   performance}.

We can clearly see that, max pooling is almost always better than average pooling except for
soft threshold with fewer pyramid levels.      This is consistent with recent results for image
classification \cite{boureau2010learning}.
From Fig.~\ref{Fig:Pooling}, we also see that soft threshold encoding achieves increasing
accuracies as pyramid depth increases, especially on the challenging LFW-a dataset \cite{LFW_a}.
As for sparse coding, the performance does not improve significantly when the pyramid
depth is already more than three.
With max pooling   and   three or more pyramid levels, soft
threshold obtains close results with sparse  coding on AR and even better results on LFW-a.
Note that soft threshold encoding is simple and computationally much more efficient than sparse coding.

\begin{table}[]
\centering
\begin{tabular}{r ccc}
\hline
\multirow{2}{*}{Test ~ ~ } & \multicolumn{3}{c}{Dictionary}\\
 \cline{2-4}
 & AR & ExYale B  & CIFAR-10\\
 \hline
 AR (ST) & 97.8 & 97.6 & 97.5\\
AR (SC) & 98.2 & 98.2 & 97.8\\
  ExYale B (ST) &99.8  & 99.8 & 99.8\\%
  ExYale B (SC) & 98.0 & 98.7 & 98.3\\%
 \hline
\end{tabular}
\caption{Recognition accuracies (\%) on AR and Extended Yale B (ExYale B) of our method using different datasets (listed on the second row) for dictionary generation. }
\label{Tab:datasets}
\end{table}

\subsection{Representing faces using a dictionary learned from non-face image patches}

In this subsection, we test whether the training data for dictionary learning is
critical. We generate the dictionary using both the training data from the testbed dataset
and on additional datasets. For example, we test on the AR dataset with dictionary trained
on AR or Extended Yale B.
From Table~\ref{Tab:datasets}, we can see that on par
results are obtained by training on the same  dataset from the test data or a different one.

In addition, we further evaluate the dictionary generated
on datasets of {\em non-face} images. Here we use a subset of
CIFAR-10\footnote{\url{http://www.cs.toronto.edu/~kriz/cifar.html}}, which includes 10,000
natural images.
It is surprising that learning a dictionary using the dataset of non-face images also works very
well.
\emph{That is, a face image can be recognised using a representation learned from
non-face images.} This is impossible for the representation error based methods like SRC.
This property is desirable because one can obtain a dictionary that is as large as needed
even when lacking of training face data.

   The reason why learning face representations using non-face data works  so well
   might be as follows.   We have used the single-layer unsupervised feature learning
   to generate the dictionary. It is well known that single-layer dictionary learning
   such as sparse coding mainly learns image filters that resemble Gabor-like filters.
   Small patches extracted from either face images or non-face images would both learn
   Gabor-like filters.
   However, for multiple-layer hierarchical feature learning, one expects that using
   object-specific training data might learn more meaningful high-level features.

\section{Experimental results}
\label{SEC:Exp}

In this section, we compare our methods with several state-of-the-art algorithms on
four benchmark face recognition databases. These compared algorithms include SRC, CRC,
robust sparse coding (RSC \cite{RSC11}), the Fisher discrimination criterion based dictionary
learning method FDDL \cite{yang2011fisher}, patched based MSPCRC \cite{zhu2012multi}
and PNN \cite{kumar2011maximizing}. For these methods we use the codes and recommended
settings provided by the authors.

From the pilot experiments in the last section, we know that the important factors for the final
face recognition performance are: the image patch size and stride step, encoding methods, and the pooling strategy.
We now report results using various encoding methods; and the patch size and stride step are set to be small
      ($6\times 6$ pixels and the stride step is 1 pixel in all of the experiments).
All the results of our method in this section are based
on max-pooling with 5 pyramid pooling levels.
We set the dictionary size to 1,600 in all experiments, which are randomly sampled from 50,000 small patches.
Therefore,
   no learning is involved in dictionary learning for our approach.

\subsection{Extended Yale B}
The Extended Yale B dataset \cite{GeBeKr01} consists of 2414 frontal face images from 38
subjects under various lighting conditions. The images are cropped and normalized to $192
\times 168$ pixels [31].
To increase the difficulty, in each class we randomly choose only 10 sample for
training and another 5 samples for testing. All the images are finally resized to $32
\times 32$. We run experiments on 5 independent data splits and report the average results in
Table~\ref{Tab:Yale}.

It is clear that our method with soft threshold encoding achieves the highest average accuracy of
97.3\%, which is  better than MSPCRC. In contrast, the best result among all other
methods is only 92.6\% (obtained by FDDL). Our method with sparse coding does
not perform as
well as soft threshold on this dataset. We have not carefully tuned the regularization parameter in sparse coding.
Careful tuning of this parameter might improve the performance of sparse coding.

We also list  in Table~\ref{Tab:Yale2} the recent published results of several sate-of-the-art methods on this dataset. It is clear that our method outperforms all other methods with 99.8\% accuracy.
{\em To  our knowledge, ours is the best published result with the same settings on the Extended Yale B dataset.
}

\begin{table*}[]
\centering
\begin{tabular}{r  ccccccc}
\hline
Method & CRC & SRC & RSC &FDDL & MSPCRC & Ours (ST)&Ours (SC)
\\   \hline
Accuracy&$91.9 \pm 0.8$&$89.0 \pm 1.2$&$89.7 \pm0.5$&$92.6\pm 1.3$&$97.1 \pm 1.5$&${\bf 97.3 \pm 1.3}$&$91.0 \pm 2.1$\\
 \hline
\end{tabular}
\caption{Recognition accuracies (\%) and their standard deviations on the Extended Yale B dataset with dimension $32 \times 32$. Results are based on 5 independent runs.
 }
\label{Tab:Yale}
\end{table*}

\begin{table}[h]
\begin{tabular}{r | ccc}
\hline
Method & Training size & Dimension & Accuracy\\
\hline\hline
DLRD \cite{ma2012sparse} & 32 & $24 \times 21$ & 98.2\\
FDDL \cite{yang2011fisher} & 20 & $54 \times 48$ & 91.9\\
LC-KSVD \cite{jiang2011learning} &32 & $24 \times 21$ & 96.7\\
LSC \cite{theodorakopoulos2011face} & 32 & $104 \times 91$ & 99.4\\
Low-Rank \cite{chen2012low} & 32 & 100 (PCA) & 97.0\\
RSC \cite{RSC11} & 32 & 300 (PCA) & 99.4\\
Ours & 32 & $32 \times 32$ &\textbf{99.8}\\
\hline
\end{tabular}
\caption{Published results (\%) in the literature on the Extended Yale B dataset. Number of training images
per class are shown in the second column and the rest are for test.
Soft threshold encoding is used for our method.
 }
\label{Tab:Yale2}
\end{table}

\begin{table*}[]
\centering
\begin{tabular}{r ccccccc}
\hline
Method & CRC & SRC & RSC &FDDL & MSPCRC & Ours (ST)&Ours (SC)\\
          \hline
$32 \times 32$&$59.4 \pm 1.1$&$75.5 \pm 1.5$&$73.0 \pm 2.2$&$79.4\pm 1.4$&$41.0 \pm 1.6$&${\bf 94.3 \pm 0.5}$&${\bf 95.9 \pm 1.0}$\\
$64 \times 64$&$54.0 \pm 0.8$ & $73.6 \pm 1.5$&$73.0 \pm 2.2$ & $79.5 \pm 1.8$ &$41.9 \pm 2.2 $ & ${\bf 97.0 \pm 0.7}$ & ${\bf98.8 \pm 0.5}$\\
 \hline
\end{tabular}
\caption{Recognition accuracies (\%) and their standard deviations on FERET with dimension $32 \times 32$ and $64 \times 64$. Results are based on 5 independent runs.}
\label{Tab:FERET}
\end{table*}
\subsection{FERET}
The FERET dataset \cite{FERET} is another widely-used standard face recognition benchmark set
provided by DARPA.
We use a subset of FERET which includes 200 subjects.
Each individual contains
7 samples, exhibiting
 facial expressions, illumination changes and up to 25 degrees of pose variations. It is
composed of the images whose names are marked with `ba', `bj', `bk', `be', `bf', `bd' and
`bg'. The images are cropped and resized to $80 \times 80$ \cite{yang2003combined}. We
randomly choose 5 samples in each class for training and the rest 2 samples for testing.
The mean results of 5 independent runs with down-sampled $32 \times 32$ and $64 \times 64$
images are reported in Table~\ref{Tab:FERET}.

Different from Extended Yale B, FERET has a large number of classes and exhibits pose variations.
On this dataset, we can see that our methods  perform significantly better than previously reported results.
   Almost all other methods do not achieve satisfactory results.

{\em Our method with sparse coding obtains an average accuracy of 98.8\% with image dimension $64
\times 64$ pixels, which outperforms the previous best result of FDDL \cite{yang2011fisher}
by a large margin of about 20\%.}
Note that our method is based on randomly extracted
patches, while the dictionary of FDDL is trained by a very expensive process. MSPCRC does
not perform well on this dataset.

\subsection{AR}

\begin{table*}[]
\centering
{
\begin{tabular}{r cccccc}
\hline
Method & CRC & SRC & RSC &FDDL  & Ours (ST) &Ours (SC) \\
\hline
AR (Occlusion)& 84.5&86.5&88.5&83.5& \textbf{97.8}&\textbf{98.2} \\
AR (Clean) &88.4&92.9&92.3&85.9&\textbf{98.4}&\textbf{98.9}
\\
 \hline
\end{tabular}
}
\caption{Recognition accuracies (\%) on the AR dataset with dimension of $32 \times 32$ pixels. ST and SC are our methods with soft threshold and sparse coding, respectively.}
\label{Tab:AR}
\end{table*}

\begin{table}
\centering
\begin{tabular}{r | ccc}
\hline
Method & Situation & Dimension & Accuracy\\
\hline\hline
LSC \cite{theodorakopoulos2011face} &Clean &$116 \times 84$&90.0\\
 FDDL \cite{yang2011fisher} & Clean&$60 \times 43$&92.0\\
DLRD \cite{ma2012sparse}& Clean & $27 \times 20$ & 95.0\\
RSC \cite{RSC11} & Clean & 300 (PCA) & 96.0\\
Ours & Clean & $64 \times 64$ & \textbf{99.1}\\
L2 \cite{Javen2011} & Occlusion & $168 \times 120$&95.9\\
ESRC \cite{ESRC2012} & Occlusion &$168 \times 120$&97.4\\
SSRC \cite{dengdefense2013} & Occlusion&$168 \times 120$ & 98.6\\
Ours  & Occlusion & $64 \times 64$ & \textbf{99.2}\\
\hline
\end{tabular}
\caption{Published results (\%) on AR dataset in two situations. Sparse coding is used for our method.}
\label{Tab:AR2}
\end{table}
The AR dataset \cite{AMM98} consists of over 4000 facial images from 126 subjects (70 men
and 56 women). For each subject 26 facial images were taken in two separate sessions.
The images exhibit a number of variations including various facial expressions (neutral,
smile, anger, and scream), illuminations (left light on, right light on and all side
lights on), and occlusion by sunglasses and scarves. Of the 126 subjects available 100
have been selected in our experiments (50 males and 50 females). All the images are
finally resized to $32 \times 32$ pixels.

We first test on the AR dataset in two situations: with occlusions and without occlusions. For
the first situation, the 13 samples in the first session are used for training and the
other 13 images in the second session for test. For the second `clean' situation, the
7 samples in each session with only illumination and expression changes are used. From
Table~\ref{Tab:AR}, we can see that {\em both of our two methods with  soft threshold encoding
and sparse coding encoding obtain the best results on these two situations}. The previous best results of other
methods in the occlusion and clean situation are obtained by RSC and SRC, which are worse
than our method with sparse coding by 9.8\% and 10\%, respectively.

We also list in Table~\ref{Tab:AR2} the recent published results of several
sate-of-the-art methods on AR. It is clear that our method achieves the highest accuracies
(99.1\% and 99.2\% for the clean and occlusion situation, respectively). To the best of
our knowledge, these are the best results obtained on AR in these two situations.

\begin{figure}[]
\centering
\includegraphics[width=0.48\textwidth]{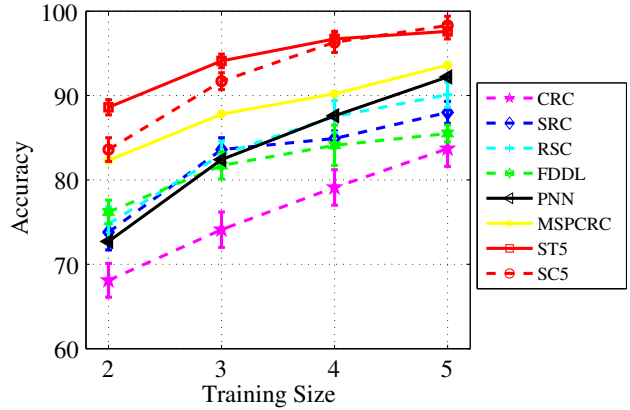}
\caption{Aveage recognition accuracies (\%) of 5 independent runs on AR dataset with number of training samples per class from 2 to 5. Standard deviations are shown with error bars. Results for PNN and MSPCRC are cited from \cite{zhu2012multi} with the same setting, where the deviation results are not shown here due to their large values.}
\label{Fig:AR_trn_sz}
\end{figure}

We further test these methods with less number of training samples per class. Among the samples
with only illumination and expression changes in each class, we randomly select 2 to 5
images of the first session for training and 3 of the second session for test. We run
the experiments 5 times and report the average results in Fig.~\ref{Fig:AR_trn_sz}. It
is clear that, our methods consistently outperform all other methods by large margins in
all situations. An accuracy of 88\% is obtained by soft threshold with only 2 training
samples per class. With less training samples, soft threshold performs better than sparse
coding on this dataset. Holistic image based method CRC achieves the worst results, while
multi-scale patch based CRC performs the second best.
This demonstrates that  relatively smaller-size local patches might encode more discriminative information,
which helps classification, than bigger-size patches or the whole image.

\subsection{LFW-a}
The original LFW database \cite{LFWTech} contains images of 5,749 subjects in unconstrained environment.
With the same settings as in \cite{zhu2012multi},
     here we use LFW-a, an aligned version of LFW using commercial face alignment software \cite{LFW_a}.
     A subset of LFW-a including 158 subjects with each subject more than 10 samples are used. The images are cropped to $121 \times 121$ pixels.
Following the settings in \cite{zhu2012multi},
          2 to 5 samples are randomly selected for training and another 2 samples for
testing. All the images are finally resized to $32 \times 32$.

\begin{figure}[]
\centering
\includegraphics[width=0.48\textwidth]{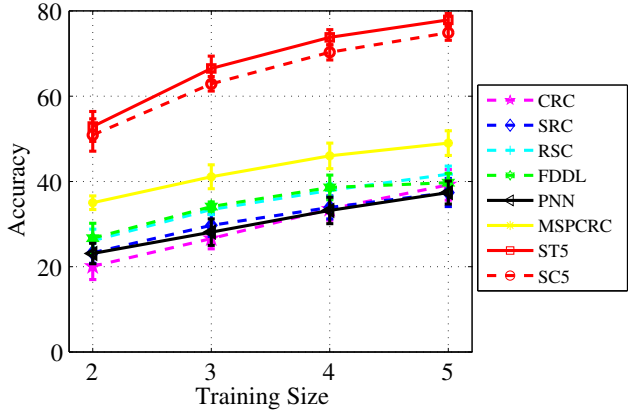}
\caption{Average recognition accuracies (\%) of 5 independent runs on the LFW-a dataset.
Standard deviations are shown with error bars. Results for PNN and MSPCRC are cited from
[3] with the same setting.}
\label{Fig:LFW_trn_sz}
\end{figure}

Fig.~\ref{Fig:LFW_trn_sz} lists  the average results of 5 independent runs. Similar results
are obtained as in Fig.~\ref{Fig:AR_trn_sz}. Our methods perform considerably better than all other
methods with even larger gaps on this challenging dataset. Same as on the AR dataset, soft threshold encoding
achieves slightly better results than sparse coding here.
 On this dataset, residual based methods (even MSPCRC and PNN that use local patches) obtain poor results.
{\em With 5 training samples each class, our approach with
 soft threshold encoding outperforms the state-of-the-art method MSPCRC by around 29\%.
}

\subsection{Feature combination}
All the previous evaluations  are based on raw intensity.
To achieve better performance, multiple complementary features can be extracted for input of the feature learning pipeline.
In this experiment, we take local binary patterns (LBP) for example to test the performance by feature fusion.
We first represent each image as an LBP code image, and then apply the feature learning pipeline  on images of raw pixels and LBP codes  independently. The learned feature vectors through the two channels for each image are then concatenated  into one large vector.

Table~\ref{Tab:Fusion} shows the compared results between raw intensity and its fusion
with LBP. Soft threshold is used for feature encoding. On both AR and LFW-a, the fused
features obtain improved accuracies with all training sizes. For example with 2 training
samples each class on LFW-a, the accuracy is averagely increased by 1.7\% with the
combined feature.
It is an interesting research topic to explore how to improve the classification by combining
more low-level image statistics.
We leave this as a future research topic.

\begin{table*}[]
\centering
\begin{tabular}{r | cccccc}
\hline
 Training size & 2 & 3 & 4 & 5\\
 \hline\hline
AR (Raw) & $88.6 \pm 0.9$ & $94.1 \pm 0.8$& $96.7 \pm 0.9$& $97.6 \pm 0.9$   & \\
AR (Raw + LBP) & $ 89.2 \pm 2.4$  &$ 95.1 \pm 1.1$ & $ 97.9 \pm 1.0$ & $ 98.3 \pm 0.6$
   & \\
  \hline
 LFW-a (Raw) & $52.9\pm3.5$  &$66.5\pm 2.9$ &$73.8 \pm 1.8$  &$77.9 \pm 1.5$ &\\
 LFW-a (Raw + LBP)&$ 54.6 \pm 2.7$& $ 66.8 \pm 1.9$&$ 74.6 \pm 2.5$& $ 78.9 \pm 1.8$\\
\hline
\end{tabular}
\caption{Average results (\%) of 5 runs on AR and LFW-a with fusion of learned features. The
first column shows the training size per class. Soft threshold is used for encoding.
}
\label{Tab:Fusion}
\end{table*}

\subsection{Modular methods vs.\ our pipeline}

In this experiment, we want to compare our approach against the modular approach which is also
based on local patches.

\begin{figure}
\includegraphics[width=0.48\textwidth]{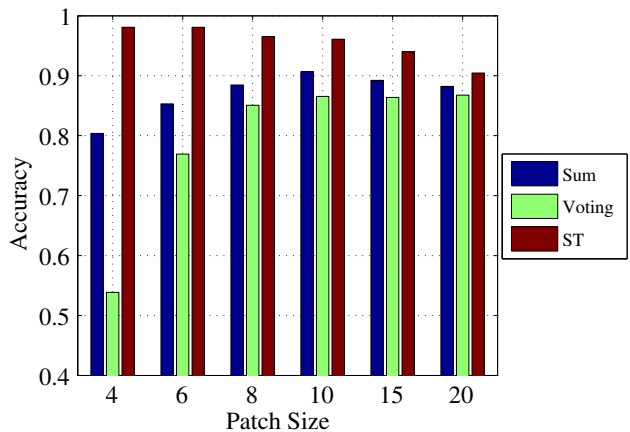}
\caption{Comparison between modular methods (with `Votin'g and `Sum' strategy)
   and our model (with soft threshold encoding) on the AR dataset.
      The patch size varies from $4 \times 4$ to $20 \times 20$ pixels.}
\label{Fig:Modular}
\end{figure}

The modular approach has been widely used for robust face recognition.
For example, the voting strategy is used to aggregate the intermediate results on local patches in
\cite{Wright09,CRCzhanglei2011}, which achieved much better results than the original methods on occlusion problems.
For efficiency, here we take CRC  \cite{CRCzhanglei2011} for example to compare the modular approach with our image classification
model on the AR dataset. Local patches are densely extracted from the images of $32 \times 32$  pixels
with a stride of 1 pixel.
The comparative results with different patch sizes are shown in Fig.~\ref{Fig:Modular}.
In addition  to voting, we also conduct the modular approach by combining the residuals of all local patches
for classification (`Sum' in the figure).
The observations are:
\begin{enumerate}
\item
Our generic image classification model achieves significantly better results than the modular approaches
with all the setting of different patch sizes.
\item
Our approach tends to achieve better results with smaller patches
while the modular approaches favour mid-size patches.
This is mainly due to the spatial pyramid pooling step in our pipeline.
It remains unclear how to introduce spatial pooling into the residual based approaches such as SRC and CRC.
\item
Also note that the simple sum operation consistently obtains higher accuracies
than the majority voting method in the modular approaches.
\end{enumerate}
   This experiment shows that besides the advantage of local patches,
   other components in our pipeline such as feature encoding and spatial pooling must have contributed to
   the superior performance of the image classification pipeline for face recognition.

\section{Conclusion}

Can face images be classified using  the same strategy as in generic image classification?
In this work, we have answered this
question by applying the generic image classification pipeline to face recognition,
which {\em outperforms all the reported state-of-the-art face recognition methods by large margins
on four standard benchmark datasets.}

A large body of recent published face recognition methods, especially SRC-scheme
based ones, have focused on designing complex dictionaries (e.g., using discriminant
learning or K-SVD etc). By a thorough cross evaluation of the components in the image classification
pipeline on face recognition problems, we show that the choice of dictionary learning algorithms
is less critical than that of encoders.  In particular, we show that randomly  selected
patches can work very well. Interestingly, we find that the dictionary is even not
necessarily generated using patches extracted from face images. That is, a face image can be recognised using a
representation generated from non-face image patches.

Another finding of this work is that the simple and efficient soft threshold algorithm
surprisingly obtains on par results with the expensive $L_1$-constrained sparse coding.
Note that, the best results on some datasets that we have tested are obtained by the
components of a dictionary of randomly selected patches, soft threshold encoding, and pooling,
followed by a simple linear classifier. In this entire procedure,
{\em the only component involving `learning' is the simple linear classifier training}.

   In the future, we plan to explore the performance of multiple-layer hierarchical feature learning for face recognition.
   Our initial experiments on learning two-layer features using the approach of \cite{KirosS12} and sparse filtering of
   \cite{SP11}
   did not improve the performance over the single-layer features of this work on the AR and LFW-a datasets.
   However, more thorough experiments are needed before we can draw a conclusion.
   The facial image descriptor extracted with the pipeline presented here has potentials for other
   faces-related applications, such as face detection, retrieval,  alignment, and tagging.
   Our preliminary experiments show very encouraging results on large-scale face image retrieval.

{
\small
\bibliographystyle{ieee}
\bibliography{face}
}

\end{document}